# On Policy Learning Robust to Irreversible Events: An Application to Robotic In-Hand Manipulation

Pietro Falco, Abdallah Attawia, Matteo Saveriano, and Donghuei Lee

*Abstract*—In this letter, we present an approach for learning in-hand manipulation skills with a low-cost, underactuated prosthetic hand in the presence of irreversible events. Our approach combines reinforcement learning based on visual perception with low-level reactive control based on tactile perception, which aims to avoid slipping. The objective of the reinforcement learning level consists not only in fulfilling the in-hand manipulation goal, but also in minimizing the intervention of the tactile reactive control. This way, the occurrence of object slipping during the learning procedure, which we consider an irreversible event, is significantly reduced. When an irreversible event occurs, the learning process is considered failed. We show the performance in two tasks, which consist in reorienting a cup and a bottle only using the fingers. The experimental results show that the proposed architecture allows reaching the goal in the Cartesian space and reduces significantly the occurrence of object slipping during the learning procedure. Moreover, without the proposed synergy between reactive control and reinforcement learning it was not possible to avoid irreversible events and, therefore, to learn the task.

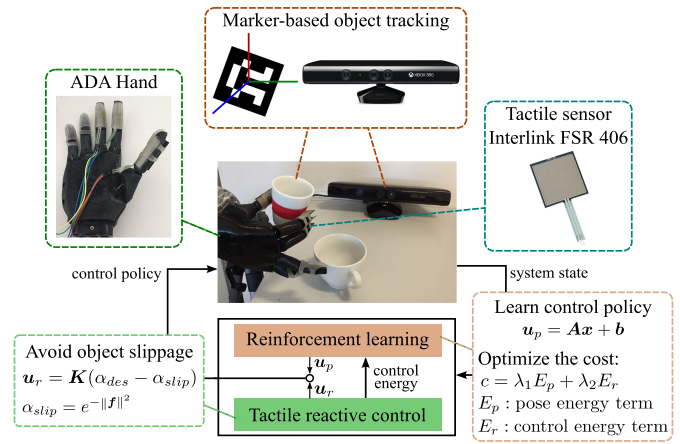

Fig. 1. Overview of the proposed architecture.

## I. INTRODUCTION

IN THE last decade, robotic systems are moving from industrial applications to service applications in human-dwelled environments. When the robots share the same environments as humans, a crucial skill is the ability to use in a straightforward fashion also tools and objects designed for humans. Hence, equipping robots with anthropomorphic hands and providing in-hand manipulation skills is a crucial step towards service robotics. When performing in-hand manipulation, the robot changes the pose of an object with respect to the palm using only the fingers. In in-hand manipulation applications, planning reliable and robust trajectories offline still remains a great challenge, since interaction forces between the object and the robotic hand are difficult to predict, especially if the robot works in unstructured environments and the object properties are not fully known. Moreover, low-cost robotics hands, which are becoming increasingly popular in the robotic community, are compliant and underactuated [1]–[4]. On one side, those hands can simplify grasping and potentially in-hand manipulation by exploiting compliance and hand synergies. On the other side, it becomes also more difficult to derive reliable mathematical models. For these reasons, reinforcement learning can play a key role in this field, as it allows learning control policies and object properties through the interaction between the robot and the environment. However, a significant limitation of strategies based on trial and error is that, in complex tasks like in-hand manipulation, *irreversible events* can occur during the learning process. For example, during a manipulation task the object may fall down and the robot is not able to easily pick it up and continue the learning procedure. As a second example, during a task involving motion planning, the robot could hit an obstacle damaging the environment or itself. Irreversible events during the learning process limit the applicability of reinforcement learning in industrial and domestic environments. In order to reduce significantly the probability of irreversible events, we introduced in previous work lower-level reactive control modules that locally correct the trajectories in position and force [5], [6]. The harmonic integration of such reactive approaches

Manuscript received September 10, 2017; accepted January 14, 2018. Date of publication January 31, 2018; date of current version February 22, 2018. This letter was recommended for publication by Associate Editor T. W. Seo and Editor W. K. Chung upon evaluation of the reviewers' comments. This work was supported in part by the Marie Sklodowska-Curie Individual Fellowship LEACON, in part by EU Project 659265, and in part by the Technical University of Munich, International Graduate School of Science and Engineering and Helmholtz Association. *(Corresponding author: Pietro Falco.)*

P. Falco was with the Department of Human-Centered Assistive Robotics, Technical University of Munich, Munich 80333, Germany. He is now with the Department of Automation Solutions Västerås, ABB Corporate Research, Forskargränd 7, Västerås 72178, Sweden (e-mail: pietro.falco@se.abb.com).

A. Attawia and M. Saveriano are with the Department of Human-Centered Assistive Robotics, Technical University of Munich, Munich 8033, Germany (e-mail: abdhallah.attawia@tum.de; matteo.saveriano@tum.de).

D. Lee is with the Department of Human-Centered Assistive Robotics, Technical University of Munich 8033, Munich Germany, and also with the Institute of Robotics and Mechatronics, German Aerospace Center (DLR), Weßling 82234, Germany (e-mail: dhlee@tum.de).

Digital Object Identifier 10.1109/LRA.2018.2800110

in a reinforcement learning framework is still an unsolved problem, especially in the field of in-hand manipulation. Within the field of reinforcement learning, the recent trend is to exploit deep learning approaches, especially to acquire object grasping skills [7], [8]. A limitation of actual deep reinforcement learning methods for in-hand manipulation is the relatively high number of interactions with the external environment required to learn the task. Hence, in real-world applications irreversible events are likely to occur, especially in the beginning of the learning process. In this work, we propose a framework to achieve in-hand manipulation avoiding irreversible events such as object slipping. Classical motion planning techniques are not easily applicable, since the model of the hand and of the object are not known. On the other hand, also sole reinforcement learning is not easily applicable, since the object can slip and pick it up again is time consuming and cannot be automated. For this reason, we propose an approach that exploits the synergy between a reactive slipping avoidance control layer and a reinforcement learning layer based on visual perception. The proposed architecture is depicted in Fig. 1. The two layers work in synergy though a bidirectional exchange of information. The Reinforcement Learning (RL) layer sends to the Reactive Control (RC) layer the trajectory associated to the current policy. The RC layer locally corrects the trajectory in order to avoid the object slipping during the learning procedure. On the other hand, the control layer sends to the learning layer information concerning the control energy during the trajectory, which quantifies how strongly the tactile reactions were needed. This way, the robot learns both to fulfill the in-hand manipulation operation and, at the same time, to avoid the intervention of instinctive reactions from the low-level reactive control. A second limitation of current (deep) reinforcement learning for in-hand manipulation is the lack of integration between different sensing modalities. State-of-the art learning approaches are mainly based on monomodal perception, mostly visual. Although such modern approaches can work for grasping tasks, they are not proven to be effective for in-hand manipulation where robot and environment interact tightly and the interaction forces play a key role. An important part of the learning-control synergy proposed in this work is the usage of higher-rate tactile perception for low-level reflexes and both visual perception via a marker tracking algorithm and tactile perception for the high-level learning.

## II. RELATED WORK

Even though robust in-hand manipulation of unknown objects is not a mature field, in general dexterous manipulation has been studied in the last decade. Many works, such as [9]–[12], use motion planning and control techniques to carry out dexterous manipulation tasks. In [13], the authors introduce a planning method for the in-hand manipulation of an object with an ellipsoid surface. Such methods assume that the physical and geometric models of the object as well as the model of the fingers are known beforehand. Hence, those algorithms cannot be used easily for solving in-hand manipulation tasks on compliant and underactuated hands. In [14], an approach is presented, which exploits tactile-based learning to manipulate an object supported by a planar surface with two robotic fingers. A ReFlex robot hand is used, which is compliant and underactuated. A method to learn finger-level manipulation skills with a low-cost hand from human demonstrations is presented in [15]. Even though this work does not show case studies concerning in-hand manipulation, the programming by demonstration approach can be interesting for initializing the learning process of in-hand manipulation skills. The authors of [16] show an approach to learn an in-hand rotation task based on adaptive optimal control. The difference with our approach is that in [16] the palm of the hand supports the object and the tactile information is not exploited. Moreover, our approach exploits reactive slipping avoidance based on a low-level closed loop that works in synergy with the reinforcement learning level. A deep learning approach combined with demonstrations is adopted in [17], to perform in-hand manipulation tasks in a simulated environment, while in [18] an in-hand non-prehensile manipulation is executed starting from human demonstration. In [19], an analytical hand model is exploited with a constrained optimization scheme, while machine learning techniques based on a random forest classifier are used to select task-specific models. A method using biomimetic active touch is shown in [20], in which a robotic finger rolls a cylinder in contact with the fingertip and in [21] a 3D-Printed Tactile Gripper is presented, able to perform cylinder re-orienting operations.

We aim at achieving in-hand manipulation task, which consists in changing the pose of an object the respect to the hand palm, only using the fingers. Even though the task may appear simple for a human, we need to tackle several problems: the number of degrees of freedom of the used robotic hand is limited compared to human hand, the object can easily slip during the learning process, and, due to the hand compliance, the position of the motor at each finger does not give sufficient information on the position of fingertip. Also, we want that the robot *learns without the continuous support of a human operator*, which means that the occurrence of irreversible events like slipping must be reduced as much as possible. Due to those challenges, a traditional learning approach based on vision is not sufficient to learn the task, since it does not prevent irreversible events during the learning phase. On the other hand, a classical control approach is not applicable in a straightforward fashion since the hand is compliant and underactuated. Hence, using a kinematic and dynamic model is not directly possible. Our research question, addressed in this work, is if a synergy between reactive control and reinforcement learning, together with multimodal visuo-tactile perception, is effective to tackle these challenges.

## III. SYSTEM SETUP

The experimental setup is constituted by an Openbionics ADA hand, a RGB-D camera for visual perception, and simple low-cost strain gauges sensors for tactile perception. In order to collect the visual data, a Kinect depth camera is used, as shown in Fig. 1. The pose of the manipulated object is computed from the Kinect camera by placing a marker on the object. To track the object pose, the ROS tool called AR Track Alvar is employed [22]. A marker similar to one shown in Fig. 2 is glued on the object and tracked using Alvar to estimate online the pose of the object. The ADA Hand shown in Fig. 1 is a 3D-printed low-cost robotic hand from Openbionics that has 5 actuated degrees of freedom. Each finger is connected to a linear motor using a tendon and all the motors are controlled by a printed circuit board (PCB) that is based around the ATMEGA 2560 microcontroller. The PCB can be programmed using the Arduino Software. Therefore, the control inputs of the hand are the positions of the linear motors that pull the tendons. The hand is provided with one tendon per finger. Tactile perception is implemented with Interlink FSR 406 sensors. An Interlink FSR 406 is a force sensitive resistor that measures the normal force applied on the manipulated object. The sensor has 39 mm square active diameter and shows a decrease in resistance with the increase of force applied on this area. It has a pressure reading range of 10 g to 10 kg. One sensor is fixed on each fingertip to compute the force with which the fingers press on the manipulated object. The pressure sensors are connected to an Arduino Uno board. In this work, the force is measured only on the thumb, the index finger, and the middle finger. The tactile sensor is chosen to have enough active area to cover the front as well as the sides of the fingers.

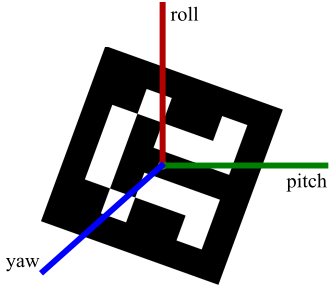

Fig. 2. Marker sticked on the object for tracking.

## IV. PROPOSED APPROACH

### A. Mathematical Description of the System

The first step of our approach consists in modeling the structure of the system. We represent the dynamic of the system as:

$$x_{t+1} = g(x_t, u_t) + w_t, \quad (1)$$

where $x_t \in X$ and $u_t \in U$ represent the state of the system and the control input at the time frame $t \in \mathbb{R}$, $g$ is the state-evolution function of the system, and $w$ is zero-mean Gaussian white noise.

In our particular application, the state of the system is constituted by the vector $x = (\phi, f) \in \mathbb{R}^4$, where $\phi \in \mathbb{R}$ is the orientation of the manipulated object described with the yaw Euler angle (see Fig. 2) and $f \in \mathbb{R}^3$ are the interaction forces at the sensorized fingertips, i.e., thumb, index, and middle finger. The control input vector $u \in \mathbb{R}^3$ includes the motor commands of the three fingers $u = (u_1, u_2, u_3)$ and the function $g$ is considered unknown. It is reasonable to consider $g$ unknown, since the hand is compliant and underactuated, the contact forces exchanged between the fingertips and the object are very difficult to predict and strongly depend on the object material.

### B. Reinforcement Learning Module

After defining a mathematical formalization for the system, we need to define a cost function in order to apply reinforcement learning. The cost is a function that maps the state space to real numbers $c : X \to \mathbb{R}$. In our approach, the cost function is defined as:

$$c(x) = \lambda_1 E_p(x) + \lambda_2 E_r(e_r(x)). \quad (2)$$

The proposed structure of $c(x)$ is constituted by two terms, denoted by the symbols $E_p$ and $E_r$. The term $E_p$ takes into account distance from the desired object configuration, and the term $E_r$, called in our framework reactive pseudoenergy, takes into account how strongly the reactive control intervenes during the task execution. Hence, $E_r$ is a function of the lower-level control error $e_r(x)$, which is in general a function of the state.

In the literature on reinforcement learning, additive terms in the cost can be used, for example, for regularization sake, in order to prevent overfitting or to limit some state variables. In this work, the meaning and the aim of the additional term is different. Our idea is that the learning algorithm leverages the reactive control module to avoid irreversible events especially in the beginning of the learning process. The terms $E_r$ brings an important benefit: it allows the system to learn, after a few iterations, also to prevent the intervention of the reactive control. The motivation of this strategy is that the reflexes are not effective in 100% of the cases. Therefore, preventing in advance the need of reactions further improves the robustness of the system. This strategy allows *a Tighter Interaction between learning and Control modules* (TIC strategy).

The complete task, then, consists in reaching the desired configuration of the object and in avoiding irreversible events such as object slipping. The terms $\lambda_1$ and $\lambda_2$ are the weights of the convex combination. We have $\lambda_2 = 1 - \lambda_1$ and $\lambda_1, \lambda_2 \in [0, 1]$. In our framework, $E_p$ and $E_r$ have the following properties:
- $E_p, E_r \in [0, 1]$.
- $E_p$ is a function of the state $x$ while $E_r$ is a function of the reactive control error $e_r$.
- $E_p = 0$ in the states where the task is fulfilled and $E_p > 0$ when the task is not fulfilled.
- $E_r = 0$ when the reactive control does not intervene and $E_r$ is a crescent function of the magnitude of the reactive control error signal.

In state of the art reinforcement learning approaches, the cost function is typically constituted only by the term $E_p$, which quantifies how well the task is executed. A classical example of $E_p$ is the distance between the final state and the desired goal. In our approach, the term $E_r$ is also crucial. We use a reactive control module to reduce the occurrence of irreversible events. However, reactions do not succeed in every case and in our approach the higher-level reinforcement learning is in charge to learn how to avoid the need of such reactions. We choose:

$$E_p = 1 - e^{-||\phi - \phi_{des}||^2}, \quad (3)$$

where $\phi_{des}$ is the desired orientation of the object. We set $\lambda_1 = \lambda_2 = 0.5$ to give equal importance to avoiding the intervention of the reactive control and to reaching the goal. In general, $E_p$ can be a function of the state $x$ according to the specific application. It is important to note that the term $E_r$ is in the cost function of the reinforcement learning module but it depends on the error in the low-level control layer. It represents an important part of the connection between the two layers. With the cost function in (3), we want to find a policy $u_p = \pi_\theta(x)$ such that (i) the robot reaches the desired state and (ii) avoids the intervention of instinctive reactions as much as possible. Due to its simplicity and effectiveness, we choose a linear policy, leaving as future work the test of more complex policies:

$$u_p = Ax + b, \qquad (4)$$

with the controller parameters being $\theta = \{A, b\}$. The goal is to find the optimal policy $\pi$ that maps from states to actions: $x \to \pi(x) = u_p$. The long-term expected return

$$J^\pi = \sum_{t=0}^{T} E[c(x_t)] \qquad (5)$$

is used in order to evaluate the performance of the learned controller $\pi$, where $c(x_t)$ is the instantaneous cost of being in state $x$ at time $t$ and $E[.]$ is the expectation operator.

In order to search for a policy that minimizes the cost function, we can adopt any reinforcement learning approach. In this work, we used PILCO (Probabilistic Inference for Learning Control) which is introduced in [23]. PILCO is a model-based policy search method that learns controllers from scratch with random initializations and without informative prior knowledge of the system. For each rollout, PILCO adopts a Gaussian Process (GP) to learn a model of the system and a gradient-based method for refining the policy. The GP, which is completely defined by a mean function $m(.)$ and a positive semidefinite covariance function $k(.,.)$, infers a distribution on the latent function $g$ in (1). The prior mean function is considered to be $m \equiv 0$ and the kernel to be the squared exponential (SE) kernel. The aim of the GP is to approximate the function $g: (x_t, u_t) \to x_{t+1}$ by using state-action pairs $(x_t, u_t)$ as training inputs and the next states $x_{t+1}$ as training target. Using the learned GP model, PILCO carries out probabilistic long-term predictions $p(x_1|\pi), ..., p(x_T|\pi)$ for a given policy $\pi$. The policy search consists of two parts: the policy evaluation and the policy improvement. This process is repeated until convergence, where an optimal policy $\pi^*$ is obtained. After the long-term predictions are computed, the expected long-term cost $J^\pi$ in (5) is evaluated by computing expected value of $c(x)$. A gradient-based policy improvement is carried out by computing the gradient $J^\pi$ with respect to the policy parameters $\theta$. Leveraging gradient-based non-convex optimization methods, the (locally) optimal policy parameter vector $\theta^*$ is computed. After the optimized policy parameters $\theta^*$ are obtained, the optimal policy is applied to the system and the new data are collected to update the GP model. In this work, we repeat this procedure for a maximum of 11 times (rollouts). For a complete description of PILCO, please refer to [23].

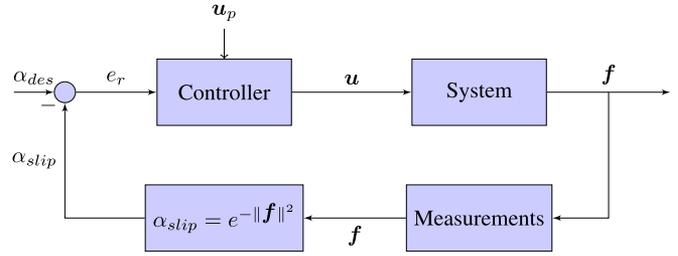

Fig. 3. Architecture of the control system.

### C. Reactive Control Module

The objective of the reactive control module is to reduce the occurrence of irreversible events, that are in this case object slipping. In order to provide effective information to the higher level, the reactive control module needs not only to detect if the irreversible event is about to occur, but it needs also to quantify the intensity of the required reaction. We call this intensity reactive pseudo-energy $E_r$. For example, if the object is slowly slipping, a slight correction of the finger motor position is needed and hence, the energy level will be close to zero. When, on the other hand, the object is slipping very fast, high level of reactive energy are required. The main task of the reactive control level is to bring the energy $E_r$ to safe levels and to communicate the higher reinforcement learning level the values of $E_r$ during the policy execution, so that it can be incorporated into the total cost.

In this particular application, the main irreversible event is the slipping of the manipulated object. To quantify the entity of the slipping, we use the following squared exponential function, which we call slipping coefficient:

$$\alpha_{slip} = e^{-\|f\|^2}, \qquad (6)$$

where $\alpha_{slip}$ has a value between 0 and 1. The value 0 indicates that the object is firmly grasped and 1 is associated to an object slippage. The vector $f$ denotes the measured forces. When the forces are near zero the slipping coefficient increases and vice versa. The chosen squared exponential satisfies the properties required to energy functions defined in Section IV-B.

The block diagram in Fig. 3 shows the reactive control system we adopted in this work. The control commands are given by the following equations:

$$u = u_p + u_r, \qquad (7)$$
$$u_r = K e_r, \qquad (8)$$
$$e_r = \alpha_{slip} - \alpha_{des}, \qquad (9)$$

where $u$ is the vector of motor commands, $u_p$ is the command from reinforcement learning level, and $u_r$ is the local correction by the reactive control module. $K \in \mathbb{R}^{3 \times 1}$ is the control gain and $e_r$ is the control error. We tuned the gain $K$ experimentally with a trial-and-error procedure. However, in problems with a more complex reactive part, $K$ could be chosen with a standard model-free control technique [24], for example Ziegler-Nichols tuning methods. Computing analytically the relationship between the slipping coefficient and the

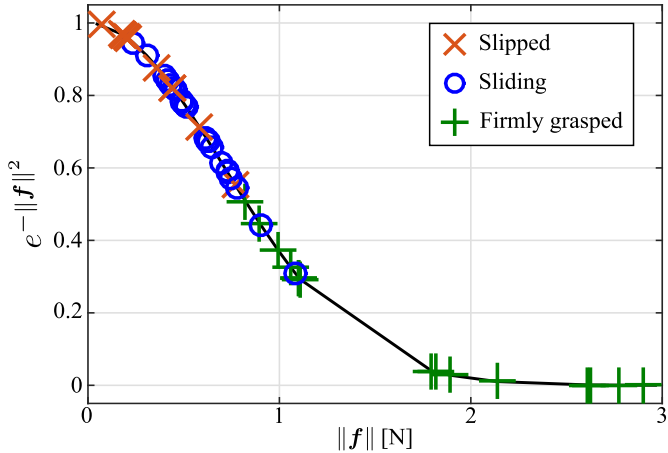

Fig. 4. Slipping coefficient as a function of the measured force norm.

TABLE I
COMPARISON BETWEEN DIFFERENT LEARNING TECHNIQUES

| Learning Technique | Success Rate | Object Slipping Rate |
|---|---|---|
| Visual RL | 0% | 100% |
| Visuo-tactile RL | 10% | 70% |
| Learning-control synergy | 90% | 0% |

## V. EXPERIMENTAL RESULTS

In this section, we evaluate our approach with the re-orienting task shown in Fig. 6. The experiment starts when the coffee cup is grasped by the hand. The motor positions for which the fingertips touch the cup are registered and used as bounds, so that during the learning procedure the finger are not open excessively. In order to show the performance of our approach, we carry out three sets of experiments. In the first set of experiments, the reinforcement learning algorithm is executed without reactive control and without using tactile data. In the second set of experiments, we include both the visual perception and tactile perception in the cost function. However, we do not activate the low-level reactive control. In the third set, we use the synergy between tactile reactive control and visual reinforcement learning. In the cost function, we include information coming from the low-level reactive control. In order to show the applicability of our approach to a different object of different material and with a larger re-orientation angle, we execute a bottle re-orienting task. For each case, 10 consecutive experiments (trials) are performed. For each experiment, we execute a maximum of 11 rollouts. The parameters of the policy $\theta$ in the first rollout are randomly initialized, as suggested in [23].

As depicted in Fig. 6, an experiment can have three possible outcomes: *task learned, task not learned*, and *object slipped*. The task is considered *learned* if (*i*) executing a maximum of 11 rollouts, the final orientation of the object $\phi \in \mathcal{G} = [\phi_{des} - 5\,\text{deg}, \phi_{des} + 5\,\text{deg}]$, (*ii*) the object does not slip for 10 s after the policy execution is over, and (*iii*) in two consecutive rollouts we have that $\phi \in \mathcal{G}$. An example of task learned is reported in Fig. 6(c). Note that a significant advantage of the proposed approach is that the task is learned directly in the operational space. We use the tolerance factor of 5 deg to take into account the error of the tracker and the limited repeatability of the object-hand system. If the object slips during the learning procedure, the outcome is *object slipped*. In Fig. 6(a) is reported an example of slipping. The object seems to reach a desired state but the hand does not hold the object firmly enough. The outcome is *task not learned* when the object does not slip, but $\phi \notin \mathcal{G}$ after 11 rollouts. An experiment in which the task is not learned is reported in Fig. 6(b), in which the object slides improperly while the hand performs the first rollouts. This is due to the fact that there is no human operator who corrects the configuration of the cup during the learning procedure. In Fig. 6(d) the learned policy is executed to pour coffee in a cup.

In order to evaluate the performance, we define two indexes (see Table I). The first is called *success rate* and is computed as the ratio between the number of successful experiments and the number of total experiments. An experiment is called successful fingertip contact forces is a very complex problem especially if the system is not in static or quasi-static condition. Therefore, we use an empirical approach. We define a function $h : \mathcal{F} \mapsto \mathcal{O}$, where $\mathcal{F} = ]0, 1]$ is the set of the forces measured during the task execution normalized between 0 and 1 through a squared exponential and, in our case, $\mathcal{O}$ is a set with three elements, i.e., $\mathcal{O} = \{\textit{firmly held, not firmly held, slipped}\}$. The class *firmly held* is associated to normalized forces such that the object does not slip. The class *not firmly held* is associated to normalized forces such that the object slowly slips, while the class *slipped* is associated to normalized forces such that the object falls down. In a preliminary calibration phase, we collected 50 force samples using random motor commands and deactivating the reactive control module. For each force sample, we label the associated class and compute the slipping coefficient $\alpha_{slip}$ according to (6). As reported in Fig. 4, we observed that for $\alpha_{slip} \in [0, 0.3)$ the object is firmly held, for $\alpha_{slip} \in [0.3, 0.7)$ the object slips within 10 seconds, and for $\alpha_{slip} \in [0.7, 1]$ the object quickly slips. Therefore, we set the desired slipping coefficient to $\alpha_{des} = 0.25$ for each finger, which is enough to hold the object without squeezing it. In higher-dimensional cases, a machine learning classification technique can be used such as support vector machines or logistic regression [25].

We define the reactive pseudoenergy as

$$E_r = |\alpha_{slip} - \alpha_{des}|. \quad (10)$$

In practice, the reactive pseudoenergy is equal to the absolute value of the control loop error defined in (9). In the literature, there are more complex slipping avoidance approaches which leverage both normal and tangential contact forces, such as [5]. Nevertheless, we use low-cost tactile sensors unable to measure the tangential component of the force. A similar approach to slipping avoidance is based on machine learning and is proposed in [26], which was specifically tested on a prosthetic hand and uses demonstrations from humans. In our case, we do not exploit human demonstrations, but we use executions from the robotic system to train our slipping model.

| Trials [#] | 1 | 2 | 3 | 4 | 5 | 6 | 7 | 8 | 9 | 10 | 11 |
|---|---|---|---|---|---|---|---|---|---|---|---|
| 1 | 1.0 | | | | | | | | | | |
| 2 | 0.94 | 0.18 | 0.11 | 0.04 | 0.07 | 0.08 | 0.17 | 0.19 | 0.13 | 0.25 | 0.19 |
| 3 | 0.92 | 0.12 | 0.18 | 0.07 | 0.15 | 0.13 | | | | | |
| 4 | 0.87 | 0.07 | 0.13 | 0.05 | 0.11 | 0.09 | 0.15 | | | | |
| 5 | 0.78 | 0.19 | 0.13 | 0.15 | 0.21 | 0.15 | 0.13 | 0.2 | 0.17 | 0.11 | 0.05 |
| 6 | 0.72 | 0.23 | | | | | | | | | |
| 7 | 0.51 | 0.11 | 0.09 | 0.09 | 0.03 | | | | | | |
| 8 | 0.51 | 0.08 | 0.09 | 0.18 | 0.11 | | | | | | |
| 9 | 0.5 | 0.04 | 0.15 | 0.03 | | | | | | | |
| 10 | 0.44 | 0.03 | | | | | | | | | |

Rollouts [#]

Fig. 5. Values of the cost function for 11 rollouts when both visual and tactile data are taken into account in the cost function. The reactive control module is deactivated. At the first rollout, the policy is initialized randomly. The green color indicates that the task is learned, blue means that the task is not learned, and red indicates that an irreversible event occurred.

when the task is learned, i.e., (i), (ii), and (iii) are satisfied. The second index is called *object slipping rate*. It is computed by the ratio between the number of experiments in which the object slips and the total number of experiments.

### A. Learning With Visual Data

In this case, the system has only the reinforcement learning level and the perception is only visual. The cost function is

$$c(\phi) = 1 - e^{-||\phi - \phi_{des}||^2}, \quad (11)$$

where $\phi_{des}$ is the desired yaw angle of the object and in this case the only state variable of the system, i.e., $x = \phi$. We specify the desired orientation of the object as $\phi = 70 \deg$. The policy is structured as $u = u_p = A\phi + b$, with $A, b \in \mathbb{R}^3$. Since the force information is totally missing and the reactive control is not active, the object slipping occurs in all the cases, as summarized in Table I.

### B. Learning With Visual and Tactile Data

In this scenario, data from the reactive control are not used, but force data are used directly in the cost function. The usage of forces in the cost function can potentially let the robot learn how to manipulate the object by maintaining suitable contact between object and fingertips. The cost function for this scenario is:

$$c(x) = 1 - (0.5a + 0.5b), \quad (12)$$
$$a = \exp(-||\phi - \phi_{des}||^2), \quad (13)$$
$$b = \exp(-||f - f_{des}||^2), \quad (14)$$

where $f_{des} = [2, 2, 2]$ N and $\phi_{des} = 70 \deg$. We performed 10 experiments. Each experiment is constituted by a maximum of 11 rollouts. The total cost values for each experiment and each rollout are plotted in Fig. 5. In this case study, the task is learned in one experiment out of ten. In two experiments, the object does not slip in none of the 11 rollouts, but the cup does not reach the final goal. In seven experiments the object slips. Even though the performance is better than in the previous case study in terms of success rate and slipping rate, this architecture achieves a success rate of 10%, which can be considered not sufficient for real-world applications. The slipping rate for this case is 70%, which means that the object slips seven times out of ten experiments. Such a rate is slightly better than in the previous case because the forces are included in the cost function. Hence, the robot tries to execute trajectories that maintain the required contact between the objects and the fingertips. However, as expected and shown in this set of experiments, using only the learning component is not sufficient to avoid slipping, especially during the first rollouts. In some experiments, even though the slipping is avoided, the cup gradually slides and the hand is not able to assign the desired object orientation, as shown, for example, in Fig. 6(b).

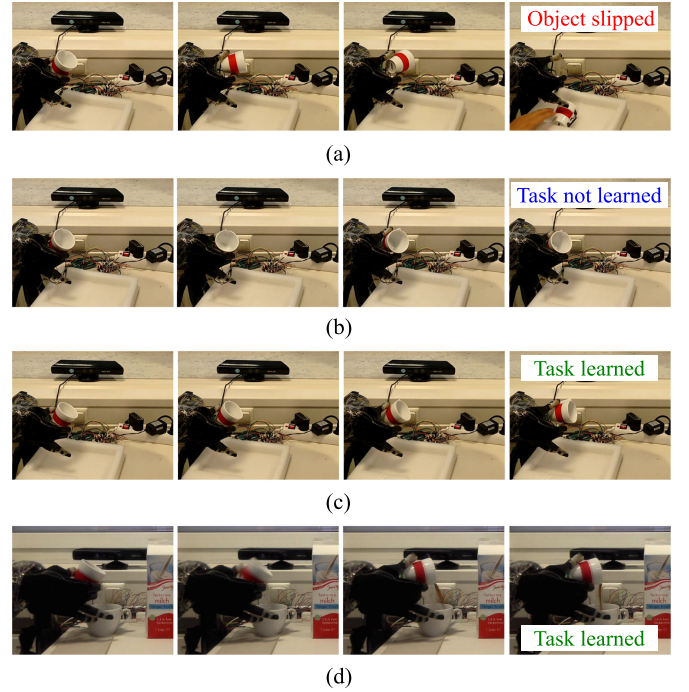

Fig. 6. Snapshots of three different experiment outcomes: (a) object slipped during the learning procedure, (b) goal not reached after 11 rollouts, (c) task learned, and (d) execution of the learned policy for pouring coffee into a cup. (a) Object slipped, (b) Task not learned, (c) Task learned, (d) Execution of the learned policy.

### C. Synergy Between Learning and Control Layers

The case in which both learning and control module work in synergy is shown in this case study. First, we analyze the performance of the proposed architecture. Then, we show how including the reactive pseudoenergy in the cost function reduces the intervention of the reactive control.

*1) Performance Analysis:* The total cost includes the goal reaching component $E_p$, that shows how far the current angle is from the target angle, as well as the reactive energy $E_r$. In particular, the cost function is the one described in (2):

$$c = \lambda_1 E_p + \lambda_2 E_r,$$
$$E_p = 1 - e^{-||\phi - \phi_{des}||^2},$$
$$E_r = |\alpha_{slip} - \alpha_{des}|,$$

| Trials [#] | 1 | 2 | 3 | 4 | 5 | 6 | 7 | 8 | 9 | 10 | 11 |
|---|---|---|---|---|---|---|---|---|---|---|---|
| 1 | 0.94 | 0.17 | 0.15 | 0.04 | 0.07 | 0.17 | 0.04 | 0.02 | | | |
| 2 | 0.92 | 0.17 | 0.04 | 0.02 | | | | | | | |
| 3 | 0.78 | 0.07 | 0.05 | 0.04 | 0.02 | | | | | | |
| 4 | 0.75 | 0.04 | 0.04 | 0.03 | | | | | | | |
| 5 | 0.75 | 0.17 | 0.04 | 0.03 | | | | | | | |
| 6 | 0.69 | 0.08 | 0.05 | 0.02 | | | | | | | |
| 7 | 0.6 | 0.14 | 0.06 | 0.15 | 0.01 | | | | | | |
| 8 | 0.56 | 0.1 | 0.17 | 0.16 | 0.14 | 0.15 | 0.04 | 0.02 | | | |
| 9 | 0.5 | 0.12 | 0.1 | 0.1 | 0.05 | 0.03 | | | | | |
| 10 | 0.49 | 0.2 | 0.09 | 0.15 | 0.16 | 0.1 | 0.18 | 0.17 | 0.16 | 0.09 | 0.19 |

Rollouts [#]

Fig. 7. Values of the cost function for 11 rollouts when the synergy between learning and control module (TIC architecture) is adopted. At the first rollout, the policy is initialized randomly. The green color indicates that the task is learned, and blue that the task is not learned.

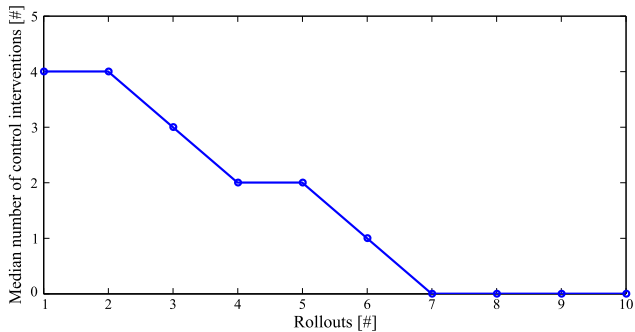

Fig. 8. Median of the number of times the control system for slipping avoidance is used in each iteration.

where $\lambda_1 = \lambda_2 = 0.5$. We use the linear policy described in (4). Differently from the case studies in Sections V-A and V-B, the control input to the robot motors $u$ is given according to a combination of the learned policies and local corrections. The results of this case study are shown in Fig. 7. The object slipping during the learning process never occurs in the performed experiments. The task is learned successfully 9 times out of 10. In one experiment, the robot performs 11 rollouts but the goal is not reached yet. Hence, the is task not learned in this case. As summarized in Table I, the proposed architecture achieves a 90% success rate. An example of orientation trajectory when executing the learned policy is shown in Fig. 10(a).

*2) Intervention of Reactive Control:* An additional set of five experiments is taken to investigate if the robot learns to prevent low-level reactions. The results are show in Fig. 8. The control reactions are considered activated when $E_r > 0.25$. The figure shows the median number of times in which the control loop intervenes to achieve a firm grasp. As we can see from the diagram in Fig. 8, in the first rollouts we have a significant number of reactions, which are needed to avoid the slipping. This explains why the architectures in Sections V-A and V-B are affected by a high object slipping rate. It is interesting to note that the median number of low-level reactions decreases to 0 after few rollouts. This means that, as expected, the system learns to avoid in advance configurations associated to unstable object grasping. This behavior is due to the presence of the control energy in the cost function.

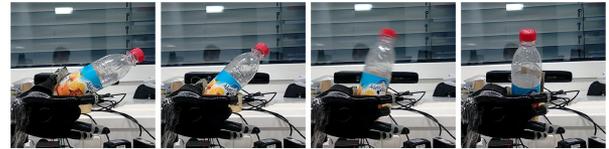

Fig. 9. Snapshot of the bottle re-orienting task.

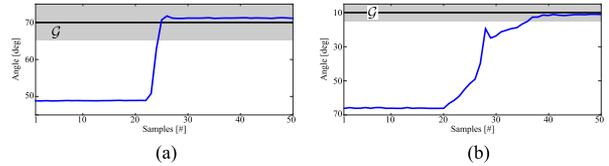

Fig. 10. Orientation of the cup (a) and the bottle (b) during the execution of the learned policy. (a) Cup Orientation, (b) Bottle Orientation.

| Trials [#] | 1 | 2 | 3 | 4 | 5 | 6 | 7 | 8 | 9 | 10 | 11 |
|---|---|---|---|---|---|---|---|---|---|---|---|
| 1 | 0.92 | 0.1 | 0.12 | 0.09 | 0.12 | 0.11 | 0.03 | | | | |
| 2 | 0.85 | 0.14 | 0.11 | 0.12 | 0.08 | 0.07 | 0.07 | 0.03 | | | |
| 3 | 0.78 | 0.11 | 0.06 | 0.04 | 0.02 | | | | | | |
| 4 | 0.72 | 0.13 | 0.09 | 0.1 | 0.08 | 0.03 | | | | | |
| 5 | 0.7 | 0.08 | 0.05 | 0.03 | | | | | | | |
| 6 | 0.69 | 0.24 | 0.23 | 0.11 | 0.11 | 0.1 | 0.1 | 0.08 | 0.09 | 0.05 | 0.06 |
| 7 | 0.67 | 0.15 | 0.08 | 0.07 | 0.03 | | | | | | |
| 8 | 0.65 | 0.09 | 0.08 | 0.06 | 0.05 | 0.05 | 0.07 | 0.02 | | | |
| 9 | 0.6 | 0.08 | 0.09 | 0.06 | 0.06 | | | | | | |
| 10 | 0.56 | 0.19 | 0.09 | 0.04 | 0.1 | 0.07 | 0.06 | 0.02 | | | |

Rollouts [#]

Fig. 11. In the experiment of in-hand re-orienting of a plastic bottle, values of the cost function for 11 rollouts when the synergy between learning and control module (TIC) is adopted. At the first rollout, the policy is initialized randomly. The green color indicates that the task is learned, blue means that the task is not learned, and red indicates that an irreversible event occurred.

### D. Experiments With a Different Object

In order to validate our approach with a different object and a different material, we performed a second experiment with the system shown in Fig. 9. In this setup, the robotic hand re-orients a plastic bottle. The material is more slippy and a desired re-orientation angle is bigger with respect to the previous task. In this case, the initial orientation of the bottle is around 70 deg and the desired goal is $\phi_{des} = 10$ deg. The performance, in terms of success rate, is summarized in Fig. 11. Also in this experiment, we perform 10 learning experiments. In each experiment, we perform 11 rollouts starting with a random policy. If the task is not learned after 11 rollouts, we consider the learning procedure not successful. If the object slips, we assume that an irreversible event has occurred and the learning procedure is not successful. As shown in Fig. 11, the task is successfully learned in 80% of the cases. An irreversible event (object slipping) occurs in one case. The task is not learned after 11 rollouts in 10% of the cases, i.e., in one experiment the final desired orientation is not in the interval $\mathcal{G} = [\phi_{des} - 5\,\text{deg}, \phi_{des} + 5\,\text{deg}]$ after 11 rollouts. An example of orientation trajectory when executing the learned policy is shown in Fig. 10(b).

## VI. CONCLUSION AND FUTURE WORK

We proposed an approach for the harmonic combination of tactile reactive control and visual reinforcement learning. The reinforcement learning module aims not only at reaching a goal in the operational space, but also at minimizing the intervention of the reactive control. Since reactive control cannot success in all the cases, learning to avoid the need of reactions can improve the performance. The experiments show that with our approach it is possible for a robotic hand to learn an in-hand manipulation task avoiding critical events such as object slipping. Another interesting aspect is that without the learning-control synergy the task was learned only in 10% of cases, since it was not possible to avoid the cup slipping during the learning procedure. In fact, when we apply the classical RL strategy, irreversible events easily occur in the first rollouts, when the robot did not gain yet enough knowledge of the environment. The TIC architecture allowed the robot to improve the knowledge of the world while having a-priori, very simple reflexes that reduce irreversible events in the first rollouts. Moreover, thanks to the learning-control synergy, the system learns also how to prevent the intervention of reactive control on the long run, enhancing further the robustness. If we do not consider the object slipping as irreversible event, a human operator can put again and again the object in the robotic hand after it has slipped. In such a case, we expect that the task could be in theory learned without the synergy with the control level. The experiments show also that the success of the learning-control synergy is connected to multimodality. In fact, marker-based visual perception is a good way to estimate with low-frequency the pose of the object. Higher-rate tactile perception is not sufficient for easily estimating the object pose but it is essential to prevent slipping. We applied this framework to an in-hand manipulation task. However, this is a general framework that can be applied in tasks which involve motion and force control of autonomous systems, by defining the task pseudoenergy $E_p$ and reactive pseudoenergy $E_r$ accordingly.

A first direction for the future work is to introduce an exploration step to estimate automatically the slipping coefficient. The second direction will be to use different robotic hands to fulfill more complex in-hand manipulation tasks such as spinning a pen. We will also apply the proposed strategy to enrich our previous work on impedance learning [27] and manipulation for humanoids [28]. Moreover, we can use a residual-based learning approach [29], [30] which exploits a very rough system model in order to reduce the number of rollouts and the learning time.

## REFERENCES

[1] R. R. Ma, L. U. Odhner, and A. M. Dollar, "A modular, open-source 3d printed underactuated hand," in *Proc. IEEE Int. Conf. Robot. Autom.*, 2013, pp. 2737–2743.
[2] D. M. Aukes et al., "Design and testing of a selectively compliant underactuated hand," *Int. J. Robot. Res.*, vol. 33, no. 5, pp. 721–735, 2014.
[3] R. Deimel and O. Brock, "A novel type of compliant and underactuated robotic hand for dexterous grasping," *Int. J. Robot. Res.*, vol. 35, no. 1–3, pp. 161–185, 2016.
[4] L. U. Odhner et al., "A compliant, underactuated hand for robust manipulation," *Int. J. Robot. Res.*, vol. 33, no. 5, pp. 736–752, 2014.
[5] G. De Maria, P. Falco, C. Natale, and S. Pirozzi, "Integrated force/tactile sensing: The enabling technology for slipping detection and avoidance," in *Proc. IEEE Int. Conf. Robot. Autom.*, 2015, pp. 3883–3889.
[6] P. Falco and C. Natale, "Low-level flexible planning for mobile manipulators: A distributed perception approach," *Adv. Robot.*, vol. 28, no. 21, pp. 1431–1444, 2014.
[7] S. Levine, C. Finn, T. Darrell, and P. Abbeel, "End-to-end training of deep visuomotor policies," *J. Mach. Learn. Res.*, vol. 17, no. 39, pp. 1–40, 2016.
[8] I. Popov et al., "Data-efficient deep reinforcement learning for dexterous manipulation," arXiv:1704.03073, 2017.
[9] J. Shi, J. Z. Woodruff, P. B. Umbanhowar, and K. M. Lynch, "Dynamic in-hand sliding manipulation," *IEEE Trans. Robot.*, vol. 33, no. 4, pp. 778–795, Aug. 2017.
[10] Z. Doulgeri and L. Droukas, "On rolling contact motion by robotic fingers via prescribed performance control," in *Proc. IEEE Int. Conf. Robot. Autom.*, 2013, pp. 3976–3981.
[11] J.-A. Seon, R. Dahmouche, and M. Gauthier, "Enhance in-hand dexterous micromanipulation by exploiting adhesion forces," *IEEE Trans. Robot.*, vol. 34, no. 1, pp. 113–125, Feb. 2017.
[12] Y. Bai and C. K. Liu, "Dexterous manipulation using both palm and fingers," in *Proc. IEEE Int. Conf. Robot. Autom.*, 2014, pp. 1560–1565.
[13] K. Hertkorn, M. A. Roa, and C. Borst, "Planning in-hand object manipulation with multifingered hands considering task constraints," in *Proc. IEEE Int. Conf. Robot. Autom.*, 2013, pp. 617–624.
[14] H. van Hoof, T. Hermans, G. Neumann, and J. Peters, "Learning robot in-hand manipulation with tactile features," in *Proc. Int. Conf. Humanoid Robots*, 2015, pp. 121–127.
[15] A. Gupta, C. Eppner, S. Levine, and P. Abbeel, "Learning dexterous manipulation for a soft robotic hand from human demonstrations," in *Proc. IEEE/RSJ Int. Conf. Intell. Robots Syst.*, 2016, pp. 3786–3793.
[16] V. Kumar, E. Todorov, and S. Levine, "Optimal control with learned local models: Application to dexterous manipulation," in *Proc. IEEE Int. Conf. Robot. Autom.*, May 2016, pp. 378–383.
[17] A. Rajeswaran, V. Kumar, A. Gupta, J. Schulman, E. Todorov, and S. Levine, "Learning complex dexterous manipulation with deep reinforcement learning and demonstrations," arXiv:1709.10087, 2017.
[18] V. Kumar, A. Gupta, E. Todorov, and S. Levine, "Learning dexterous manipulation policies from experience and imitation," arXiv:1611.05095, 2016.
[19] M. V. Liarokapis and A. M. Dollar, "Learning task-specific models for dexterous, in-hand manipulation with simple, adaptive robot hands," in *Proc. IEEE/RSJ Int. Conf. Intell. Robots Systems*, 2016, pp. 2534–2541.
[20] L. Cramphorn, B. Ward-Cherrier, and N. F. Lepora, "Tactile manipulation with biomimetic active touch," in *IEEE Int. Conf. Robot. Autom.*, May 2016, pp. 123–129.
[21] B. Ward-Cherrier, N. Rojas, and N. F. Lepora, "Model-free precise in-hand manipulation with a 3D-printed tactile gripper," *IEEE Robot. Autom. Lett.*, vol. 2, no. 4, pp. 2056–2063, Oct. 2017.
[22] S. Garrido-Jurado, R. Muoz-Salinas, F. Madrid-Cuevas, and M. Marn-Jimnez, "Automatic generation and detection of highly reliable fiducial markers under occlusion," *Pattern Recognit.*, vol. 47, no. 6, pp. 2280–2292, 2014. [Online]. Available: http://www.sciencedirect.com/science/article/pii/S0031320314000235
[23] M. P. Deisenroth, D. Fox, and C. E. Rasmussen, "Gaussian processes for data-efficient learning in robotics and control," *IEEE Trans. Pattern Anal. Mach. Intell.*, vol. 37, no. 2, pp. 408–423, Feb. 2015.
[24] K. Ogata and Y. Yang, *Modern Control Engineering*. Englewood Cliffs, NJ, USA: Prentice-Hall, 1970.
[25] C. M. Bishop, *Pattern Recognition and Machine Learning (Information Science and Statistics)*. New York, NY, USA: Springer-Verlag, 2006.
[26] F. Cordella, C. Gentile, L. Zollo, R. Barone, R. Sacchetti, A. Davalli, B. Siciliano, and E. Guglielmelli, "A force-and-slippage control strategy for a poliarticulated prosthetic hand," in *Proc. IEEE Int. Conf. Robot. Autom.*, 2016, pp. 3524–3529.
[27] F. Winter, M. Saveriano, and D. Lee, "The role of coupling terms in variable impedance policies learning," in *Proc. Int. Workshop Human-Friendly Robot.*, 2016.
[28] H. Kunori, D. Lee, and Y. Nakamura, "Associating and reshaping of whole body motions for object manipulation," in *Proc. Int. Conf. Intell. Robots Syst.*, 2009, pp. 5240–5247.
[29] M. Saveriano, Y. Yin, P. Falco, and D. Lee, "Pi-rem: Policy improvement with residual model learning," in *Proc. 10th Int. Workshop Human-Friendly Robot.*, Napoli, Italy, 2017.
[30] M. Saveriano, Y. Yin, P. Falco, and D. Lee, "Data-efficient control policy search using residual dynamics learning," in *Proc. IEEE/RSJ Int. Conf. Intell. Robots Syst.*, 2017, pp. 4709–4715.